\documentclass[times,twocolumn,final,authoryear]{elsarticle}

\usepackage{prletters}
\usepackage{framed}

\usepackage{amssymb}
\usepackage{latexsym}
\usepackage{multirow}
\usepackage{stmaryrd}

\usepackage{url}

\usepackage{amsmath}
\usepackage{graphicx}
\usepackage{subfig}
\usepackage{adjustbox}
\usepackage{booktabs}
\usepackage{color}
\usepackage{ulem}
\usepackage{makecell}
\usepackage{hhline}
\usepackage[table,usenames,dvipsnames]{xcolor}

\definecolor{newcolor}{rgb}{.8,.349,.1}
\definecolor{lightgray}{gray}{0.89}
\definecolor{markcolor}{rgb}{0.0, 0.0, 0.8}

\journal{Pattern Recognition Letters}

\begin{document}

\ifpreprint
  \setcounter{page}{1}
\else
  \setcounter{page}{1}
\fi

\begin{frontmatter}

\title{HarrisZ$^+$: Harris Corner Selection for Next-Gen Image Matching Pipelines}
 
\author[1]{Fabio \snm{Bellavia}\corref{cor1}} 
\ead{fabio.bellavia@unipa.it}
\author[2]{Dmytro \snm{Mishkin}}
\cortext[cor1]{Corresponding author: 
	Tel.: +39-091-23891124}

\address[1]{Dept. of Mathematics and Computer Science, Universit\`{a} Degli Studi di Palermo, Italy}
\address[2]{Visual Recognition Group, Faculty of Electrical Engineering, CTU in Prague}

\received{TBA}
\finalform{NA}
\accepted{NA}
\availableonline{NA}
\communicated{S. Sarkar}

\begin{abstract}
Due to its role in many computer vision tasks, image matching has been subjected to an active investigation by researchers, which has lead to better and more discriminant feature descriptors and to more robust matching strategies, also thanks to the advent of the deep learning and the increased computational power of the modern hardware. Despite of these achievements, the keypoint extraction process at the base of the image matching pipeline has not seen equivalent progresses. This paper presents HarrisZ$^+$, an upgrade to the HarrisZ corner detector, optimized to synergically take advance of the recent improvements of the other steps of the image matching pipeline. HarrisZ$^+$ does not only consists of a tuning of the setup parameters, but introduces further refinements to the selection criteria delineated by HarrisZ, so providing more, yet discriminative, keypoints, which are better distributed on the image and with higher localization accuracy. The image matching pipeline including HarrisZ$^+$, together with the other modern components, obtained in different recent matching benchmarks state-of-the-art results among the classic image matching pipelines. These results are quite close to those obtained by the more recent fully deep end-to-end trainable approaches and show that there is still a proper margin of improvement that can be granted by the research in classic image matching methods.
\end{abstract}




\end{frontmatter}


\vspace{-1em}
\section{Introduction}\label{introduction}
\vspace{-0.5em}
Image keypoint extraction has played a relevant role in computer vision since the early days~\citep{szeliski_book}. A keypoint is generally defined in a broad sense as a local region on the image which can be correctly re-localized and distinguished from others after a transformation of the image, i.e. keypoints must be repeatable and discriminable. The precision and the computational efficiency of the keypoint localization, the sparseness of the keypoint distribution over the images, and the kind of image transformations the keypoint must tolerate generally depend on the application purpose~\citep{local_invariant_feature_survey}.

With the increasing success of deep learning approaches in replacing the handcrafted ones, keypoint extraction more or less implicitly has been hidden within the layers of end-to-end trainable networks~\citep{d2net,d2d,superpoint,loftr} or apparently removed~\citep{dense}. Nevertheless, keypoint detectors as standalone building blocks are still actively investigated and successfully used as the base of modern and competitive image matching pipeline for 3D reconstruction in sparse Structure-from-Motion (SfM,~\cite{colmap}) and Simultaneous Localization and Mapping (SLAM,~\cite{orbslam}).

Although besieged by deep keypoint detectors, handcrafted detectors are still able to provide state-of-the-art results in SfM and SLAM applications~\citep{imw2020}. The keypoint extraction is the first step of 3D reconstruction pipeline. A crucial aspect in devising a better image matching schema is to adapt the core detector to the recent advancements obtained on the successive steps of the pipeline. On one hand, recent local image descriptors are become robust and able to cope with higher degrees of image deformations and noise~\citep{hardnet}. On the other hand, the last matching strategies exploiting local spatial constraints~\citep{blob_dtm} and those based on the RANdom SAmple Consensus (RANSAC,~\cite{ransac}) to exploit global model constraints according to scene geometry~\citep{adalam}, have been shown to tolerate much better the presence of outlier matches. Moreover, the increased computational power offered by recent GPU allows to process in parallel more keypoints and matches in a reasonable time depending on the application. It is not a coincidence that the blob-like Difference-of-Gaussian (DoG) keypoints of the Scale Invariant Feature Transform (SIFT) detector~\citep{sift}, ``unchained'' to output more keypoints by an appropriate removal of the setup thresholds, have achieved among the best results in the previous Image Matching Challenge\footnote{\scriptsize{\url{https://www.cs.ubc.ca/research/image-matching-challenge/2020/}}} (IMC2020,~\cite{imw2020}).

This paper presents HarrisZ$^+$, an upgrade to the corner-based HarrisZ detector~\citep{harrisz} for next-gen image matching pipelines. The aim of HarrisZ$^+$ is to provide more keypoints, better localized and distributed over the image, yet characterized by high repeatability and discriminability. Unlike the case of the unchained SIFT keypoints, HarrisZ$^+$ does not only consists of a simple tuning of the setup parameters so as to obtain more keypoints as output, but introduces further refinements to the selection criteria delineated by HarrisZ. According to the last Image Matching Challenge\footnote{\scriptsize{\url{https://www.cs.ubc.ca/research/image-matching-challenge/}}} (IMC2021) and SimLocMatch Image Matching contest\footnote{\scriptsize{\url{https://simlocmatch.com/}}}, the image matching pipelines based on HarrisZ$^+$ obtain state-of-art results among the classic image matching pipelines, closely following the more recent fully deep end-to-end trainable approaches. This proves that there is still a valuable margin of improvement towards the research of classic image matching methods.

The rest of the paper is organized as follows: Sec.~\ref{related_work} presents the related work, Sec.~\ref{harrisz_description} describes the proposed HarrisZ$^+$ updates after introducing the original HarrisZ detector, and Sec.~\ref{evaluation} reports the experimental evaluation. Conclusion and future work are outlined in Sec.~\ref{conclusions}.
\vspace{-1em}
\section{Related work}\label{related_work}
\vspace{-0.75em}
Corners and blobs are the two principal kinds of keypoints generally recognized in the literature~\citep{szeliski_book}, whose the Harris~\citep{harris} and SIFT~\citep{sift} detectors may respectively represent the most popular and effective extractors. Corners usually tend to be identified with junctions and blobs with homogeneous image regions, but in practice this distinction only holds for simple synthetic images and not for complex, natural images. The Harris detector is based on the filter response to a function of the eigenvalues of the autocorrelation matrix of the image intensity gradient. Other operators based on the autocorrelation matrix eigenvalues have been proposed to extract corners~\citep{forstner,shitomasi}, but the one defining the Harris corner detector is maybe the most commonly employed. The Hessian matrix has been also used to design keypoint detectors according to its determinant~\citep{beaudet}. Differently from the Harris corner detector, the SIFT detector is based on the filter response to DoG, as approximation of the Laplacian of Gaussian (LoG), which corresponds to the trace of the Hessian matrix. Filter-based detectors are generally invariant to illumination changes since keypoints are selected as local maxima of the filter response. Nevertheless, in practice, an absolute threshold on the filter response is generally set to suppress the detection of spurious noise as keypoints, which can decrease detector capability in case of low-contrast images. Pre-processing the input image by contrast enhancement techniques can be helpful in this situation~\citep{contrast_det}, notwithstanding that noise gets enhanced too. Concerning the robustness to geometric image transformations, in their original definition, detectors are generally only rotational invariant, while the scale depends on the filter windows size. Multi-scale approaches have been devised to solve scale issues, as well as to improve the robustness of the detector to noise~\citep{hess_lapl_affine}.
The Gaussian scale-space~\citep{linbook} is generally employed for this aim, yet non-linear scale-space definitions have been successfully applied too, as for the KAZE features~\citep{kaze}. The keypoint scale is generally selected according to the maximum filter response among the scales, where the Laplacian filter is a common choice. Additional robustness can be required in case of severe viewpoint changes, which is generally achieved by upgrading the scale and orientation associated with the keypoint to a local affine transformation approximating the unknown original perspective one. The affine transformation can be obtained locally and iteratively~\citep{hess_lapl_affine}, explicitly computing global synthetic warps of the image~\citep{asift}, but also by implicitly defining the keypoint filter response as a tensor~\citep{harris_tensor}. This geometric information characterizes the region around the keypoint and is used to generate the local image patch according to a reference system upon which to compute the associated keypoint descriptor. Recent state-of-the-art solutions based on deep learning such as AffNet~\citep{affnet} only require to provide the keypoint scale and location to get affine-normalized patches.

\vspace{-0.25em}
Fast and efficient keypoint extraction may be required by real-time applications on when processing a huge amount of images. One of the two main approaches in this sense is to discretize the kernel to allow a fast computation of the filter response thanks to the integral images. This solution was originally employed by the Speeded-Up Robust Feature (SURF) detector~\citep{surf} as alternative to SIFT. The other solution is to characterize the keypoint according to a comparison of the keypoint central pixel with respect to the other pixels in the local window. The Features from Accelerated Segment Test (FAST) detector~\citep{fast} is maybe the most known solution in this sense. In addition, FAST makes use of decision trees to further speed-up the computation. Worth to be mentioned among the handcrafted keypoint detector is the Maximally Stable Extremal Region (MSER) detector~\citep{mser}, which extracts blob-like structures that remain almost stable in a region growing process.

\vspace{-0.25em}
With the partial exception of FAST, machine learning has intervened into the keypoint detector design only recently.
After some initial attempts~\citep{tilde}, excluding the deep Key.Net detector~\citep{keynet} which employs both handcrafted and learned filter layers, deep detectors only appeared as components of end-to-end image matching architectures. The structure of deep detectors generally combine in order convolutional and max pooling layers followed by non-linear activation functions, which strongly resemble the structure of filter-based handcrafted detectors, with respectively a filter definition, the local maxima selection of its response and a thresholding on them. The Learned Invariant Feature Transform (LIFT) network~\citep{lift} is the first example of a complete end-to-end deep image matching architecture. Due to its complexity, LIFT cannot be trained as a whole from scratch and the training process requires handcrafted keypoints as bootstrap. An alternative solution was proposed in SuperPoint~\citep{superpoint}, where the training process becomes self-supervised thanks to a pre-training phase with synthetic images and the homographic adaptation. Unlike LIFT which is patch-based, SuperPoint relies on a fully convolutional network to process the whole image as input. The Detect-and-Describe (D2-Net,~\cite{d2net}) and the Describe-to-Detect (D2D,~\cite{d2d}) networks are other end-to-end local feature architectures, which operate in a different way to the standard computational flow where it is the descriptor that must be adapted to the keypoint detector. Specifically, D2-Net defines dense feature maps that simultaneously serve to derive the keypoints and their descriptors, while in D2D keypoints must be adapted to the information content of the descriptors. The DIScrete Keypoints (DISK) network~\citep{disk} leverages the principles of reinforcement learning to ease the complexity of the end-to-end training caused by the sparseness of the keypoints and achieved state-of-the-art results. More recently, the integration of the keypoint spatial constraints into the network design has greatly improved the final matching output. This has been done implementing attentional graph neural networks in SuperGlue~\citep{superglue}, using coarse-to-fine schemas with the Local Feature TRasformer (LoFTR,~\cite{loftr}), or considering dense descriptor approaches not employing any sort of keypoints~\citep{dense}.

\vspace{-0.5em}
\section{HarrisZ$^+$}\label{harrisz_description}
\vspace{-0.25em}
\subsection{From HarrisZ}\label{hz_desc}
The original HarrisZ~\citep{harrisz} extracts corners from an image, represented as a matrix $I\in\mathbb{R}^{m\times n}$, at different scales of the Gaussian scale-space. Defining the horizontal and vertical gradient derivatives of $I$ as the image convolution with the central difference kernel $K=[-1\;0\;1]$, i.e.
\vspace{-0.25em}
\begin{equation}\label{dx}
    I_x=I\ast K,\qquad\text{and}\qquad I_y=I\ast K^\top
\end{equation}
\vspace{-0.25em}
where $\ast$ indicates the convolution, the scale-space derivatives are defined as~\citep{hess_lapl_affine} 
\vspace{-0.25em}
\begin{equation}
    I_{\sigma,x}=G_\sigma\ast I_x \qquad\text{and}\qquad I_{\sigma,y}=G_\sigma\ast I_y\vspace{-0.3em}
\end{equation}
where $\sigma$ denotes the differentiation scale which identifies the current working scale and $G_\sigma$ is a Gaussian kernel with mean zero and standard deviation $\sigma$. Unlike the standard Harris corner detector~\citep{harris}, before computing the autocorrelation matrix both the scale-space derivatives get enhanced by a pixel-wise multiplication according to a smoothed raw edge mask $M_\sigma$
\begin{equation}
    \vspace{-0.25em}
    E_{\sigma,x}=I_{\sigma,x}\,M_\sigma \qquad\text{and}\qquad E_{\sigma,y}=I_{\sigma,y}\,M_\sigma
\end{equation}
where $M_\sigma$ is computed from the gradient magnitude
\begin{equation}
    I_{\sigma,m}=\left(I_{\sigma,x}^2+I_{\sigma,y}^2\right)^\frac{1}{2}
\end{equation}
by globally threshold on the mean $\overline{I}_{\sigma,m}$ computed on the whole $I_{\sigma,m}$, i.e.
\begin{equation}\label{mask}
    M_\sigma=G_\sigma\ast \mathcal{V}\left(I_{\sigma,m}>\overline{I}_{\sigma,m}\right)
\end{equation}
where $\mathcal{V}(\cdot)$ is the indicator function. The enhanced derivatives decrease the noise while strengthening the edges according to the working scale $\sigma$, providing a sort of non-linear scale-space.
The autocorrelation matrix at the pixel $\mathbf{x}$ is then computed as
\begin{equation}
    \mu_{\sigma}(\mathbf{x})=\left[\begin{matrix} 
                                      G_{s\sigma}\ast E_{\sigma,x}^2(\mathbf{x}) & G_{s\sigma}\ast E_{\sigma,x}(\mathbf{x})\,E_{\sigma,y}(\mathbf{x})\\
                                   	  G_{s\sigma}\ast E_{\sigma,x}(\mathbf{x})\,E_{\sigma,y}(\mathbf{x}) & G_{s\sigma}\ast E_{\sigma,y}^2(\mathbf{x})\\ 
                                   \end{matrix}\right]
\end{equation}
where $s\sigma$ defines the integration scale, i.e. the window size of the filter, on the basis of the differentiation scale $\sigma$ and the constant $s=2^\frac{1}{2}$. Actually, in the computation of the corner filter response $H_\sigma$, only the determinant $D_\sigma$ and the trace $T_\sigma$ of $\mu_\sigma$ are involved, respectively equal to the product and the sum of the eigenvalues of $\mu_\sigma$. These can be directly computed by simple pixel-wise operations on Gaussian-smoothed maps as  
\begin{align}
    D_\sigma &= (G_{s\sigma}\ast E_{\sigma,x}^2) (G_{s\sigma}\ast E_{\sigma,y}^2) - (G_{s\sigma}\ast(E_{\sigma,x}E_{\sigma,y}))^2\\
    T_\sigma &= (G_{s\sigma}\ast E_{\sigma,x}^2)+(G_{s\sigma}\ast E_{\sigma,y}^2)
\end{align}
In the original definition given by Harris and Stephens, the corner filter response is given by
\begin{equation}
    \tilde{H}_\sigma=D_\sigma - c\,T_\sigma^2
\end{equation}
where $c$ is a user-given coefficient. The HarrisZ filter response $H_\sigma$ replaces $c$ with an implicit adaptive reformulation according to the z-score normalization $\mathcal{Z}(\cdot)$ acting globally on the whole image 
\begin{equation}
    H_\sigma=\mathcal{Z}(D_\sigma)-\mathcal{Z}(T_\sigma^2)
\end{equation}
where
\begin{equation}
    \mathcal{Z}(Q)=\frac{Q-\overline{Q}}{\varsigma_Q}
\end{equation}
being $Q$ a generic image with mean $\overline{Q}$ and standard deviation $\varsigma_Q$. According to the statistical properties of $H_\sigma$ considering the whole input image $I$, a pixel $\mathbf{x}$ can be roughly classified as a flat region if $H_\sigma(\mathbf{x})$ is almost 0, or otherwise as an edge or as a corner if $H_\sigma(\mathbf{x})$ is respectively lower or higher than 0. Moreover, a corner must lie onto the edge mask $M_\sigma$. These two conditions can be coded as the binary mask $C_\sigma$ 
\begin{equation}
    C_\sigma=\mathcal{V}(H_\sigma>0)\,\mathcal{V}(M_\sigma>0.31)    
\end{equation}
where multiplication is intended pixel-wise and the 0.31 value is introduced to take into account the current scale in the binarization of $M_\sigma$, whose formal derivation can be found in the original manuscript. The binary mask $C_\sigma$ provides the initial set of corner candidates according to the input image instead of requiring a user-defined threshold which generally needs to be adjusted according to the input, as happens for the original Harris detector. $C_\sigma$ provides a sort of implicit adaptive contrast enhancement of the input image since both $H_\sigma$ and $M_\sigma$ rely on global statistics for their definitions. Note that recent state-of-the-art deep architectures use masks to filter keypoints with the aim of removing disturbing elements of the scene such as sky or people\footnotemark[2].

Finally, corners corresponding to $C_\sigma$ are selected so that only local maxima on $H_\sigma$ will survive. Differently from the standard multi-scale approach employed for instance in the Harris-affine detector, where the convolution kernels are maintained fixed while the image is down-scaled, HarrisZ keeps the input image size fixed while increasing the kernel size, as for SURF~\citep{surf}, in order to increase the detector accuracy. The final set of keypoints for the scale $\sigma$ is selected by sorting first the $C_\sigma$ pixel locations according to the decreasing values of $H_\sigma$ and then, in order, greedily removing a candidate keypoint if its distance from the already selected keypoints is less than $\lceil 3\sigma\rceil$. This strategy can resemble in some aspect the Adaptive Non-Maximal Suppression (ANMS,~\cite{anms}). 

Keypoint localization precision is then increased with parabolic sub-pixel precision on $H_\sigma$. The entire process is repeated for different scales
\begin{equation}\label{scale_idx}
    \sigma\in\{2^\frac{i}{2}: 3\le i\le 8\}
\end{equation}
Corners are filtered one last time according to the ratio of the corresponding autocorrelation matrix eigenvalues, which must be higher than 0.75 to further exclude accidental detected edges. Finally, corners are considered altogether among the different scales and sorted first by the decreasing scale values index $i$ and then by $H_\sigma$ if the maximum number of allowed output keypoints is exceeded. 

\subsection{To HarrisZ$^+$}\label{hzplus} 
\begin{figure}[ht]
	\center
	\begin{tabular}[c]{@{}c@{\hskip 0.2em}c@{\hskip 0.2em}c@{\hskip 0.2em}c@{\hskip 0.2em}}
		\rotatebox[origin=l]{90}{\hspace{1em}HarrisZ} & \includegraphics[height=0.13\textwidth]{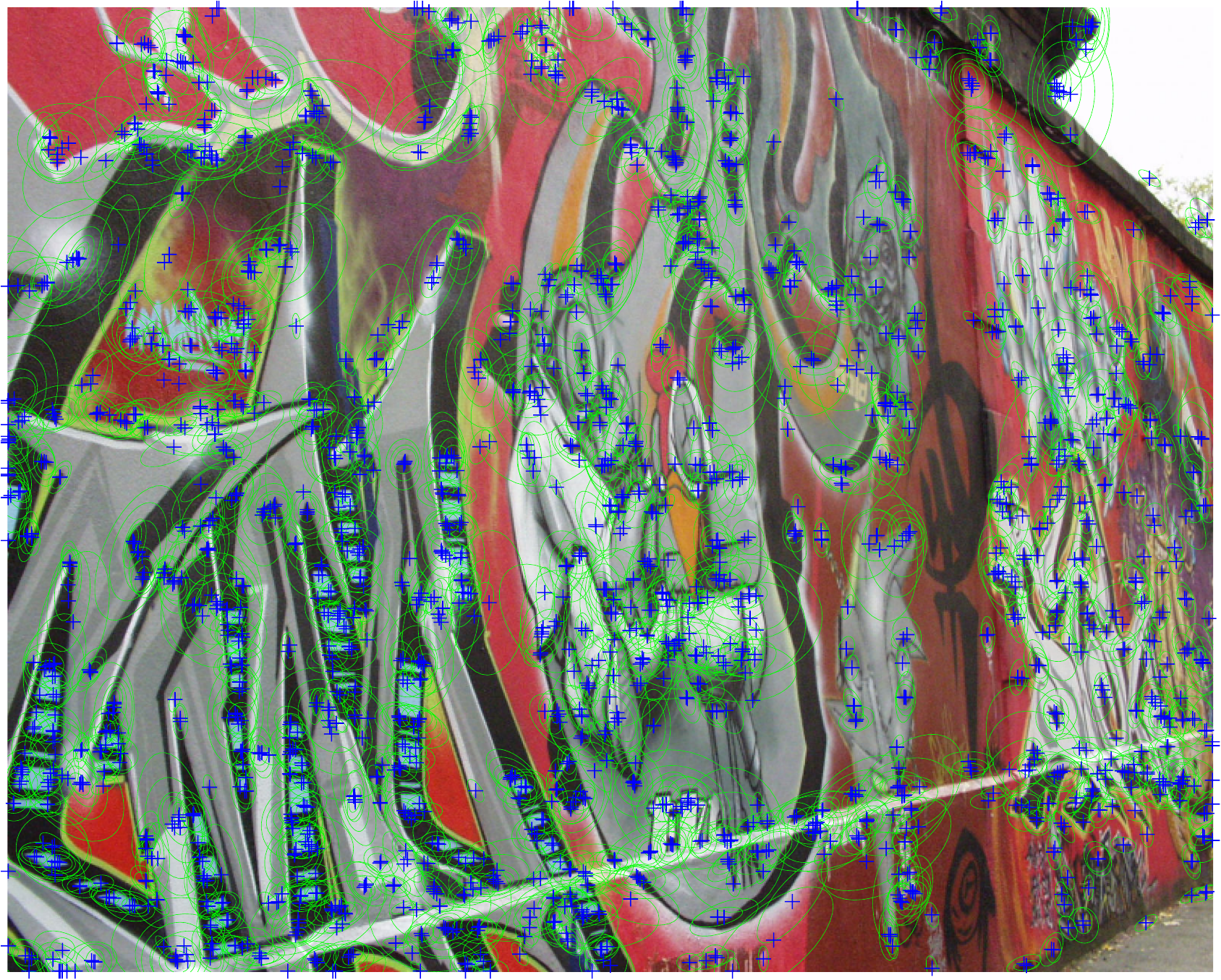} & \includegraphics[height=0.13\textwidth]{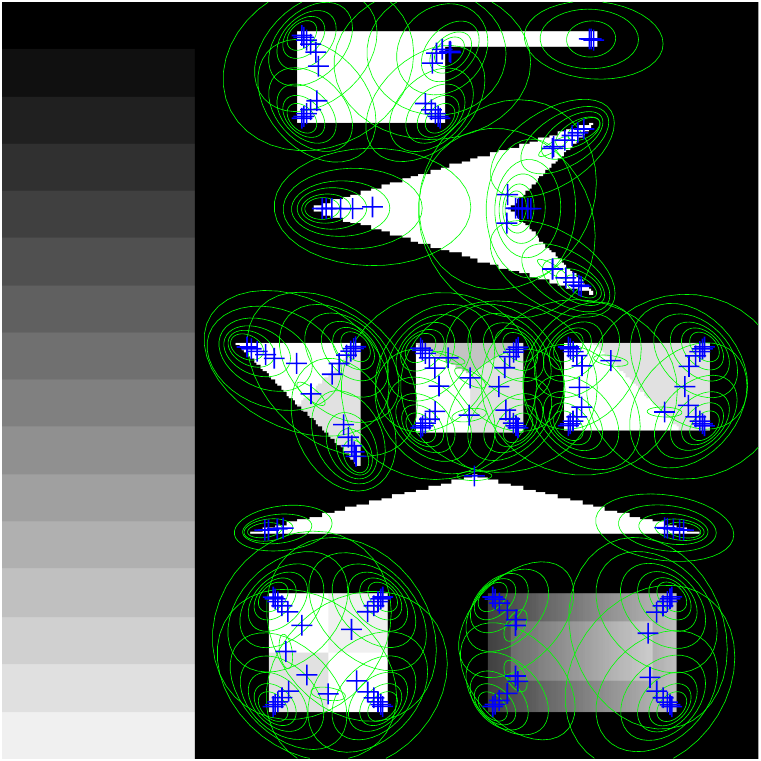} & \includegraphics[height=0.13\textwidth]{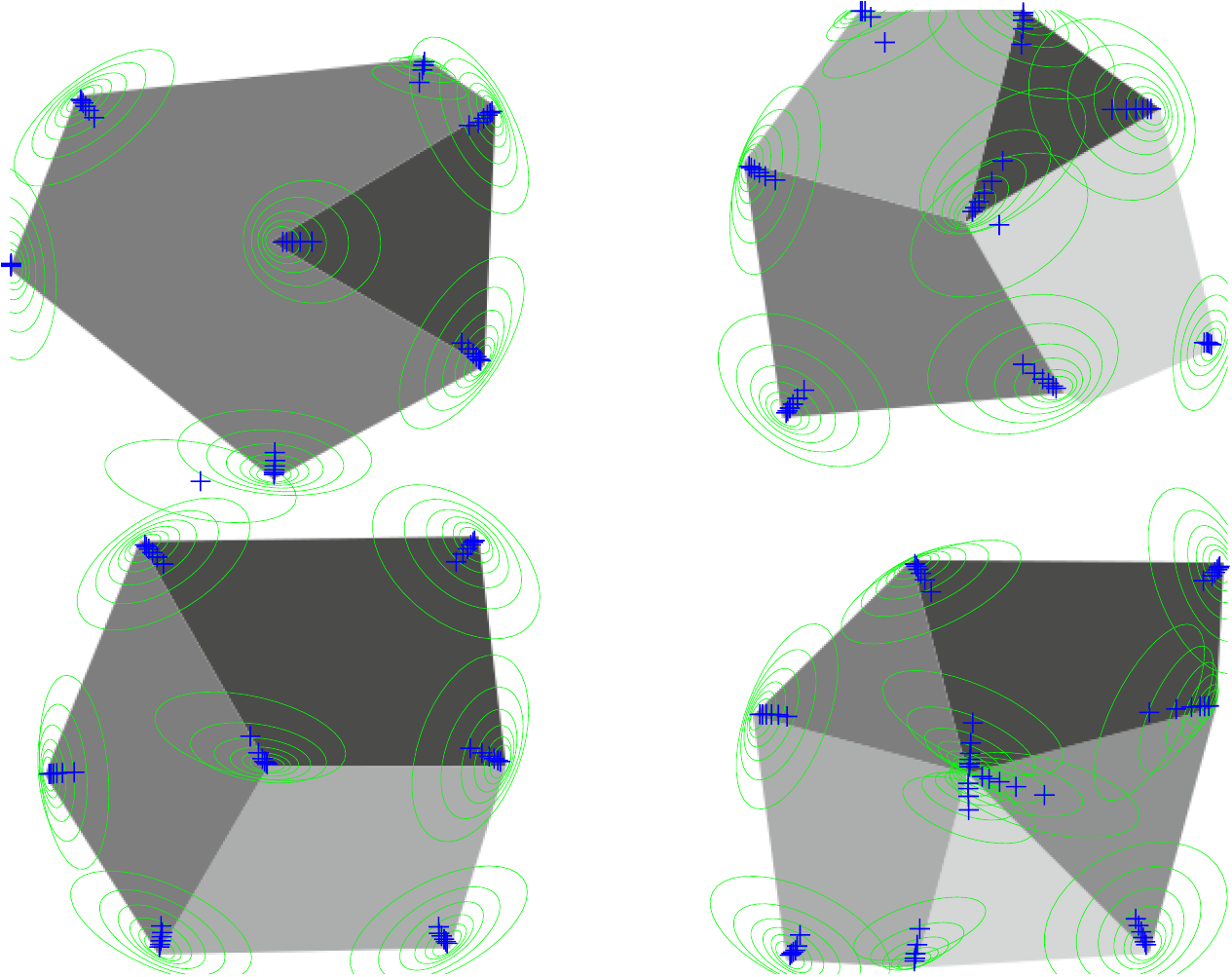}\\
		\rotatebox[origin=l]{90}{\hspace{1em}HarrisZ$^+$} & \includegraphics[height=0.13\textwidth]{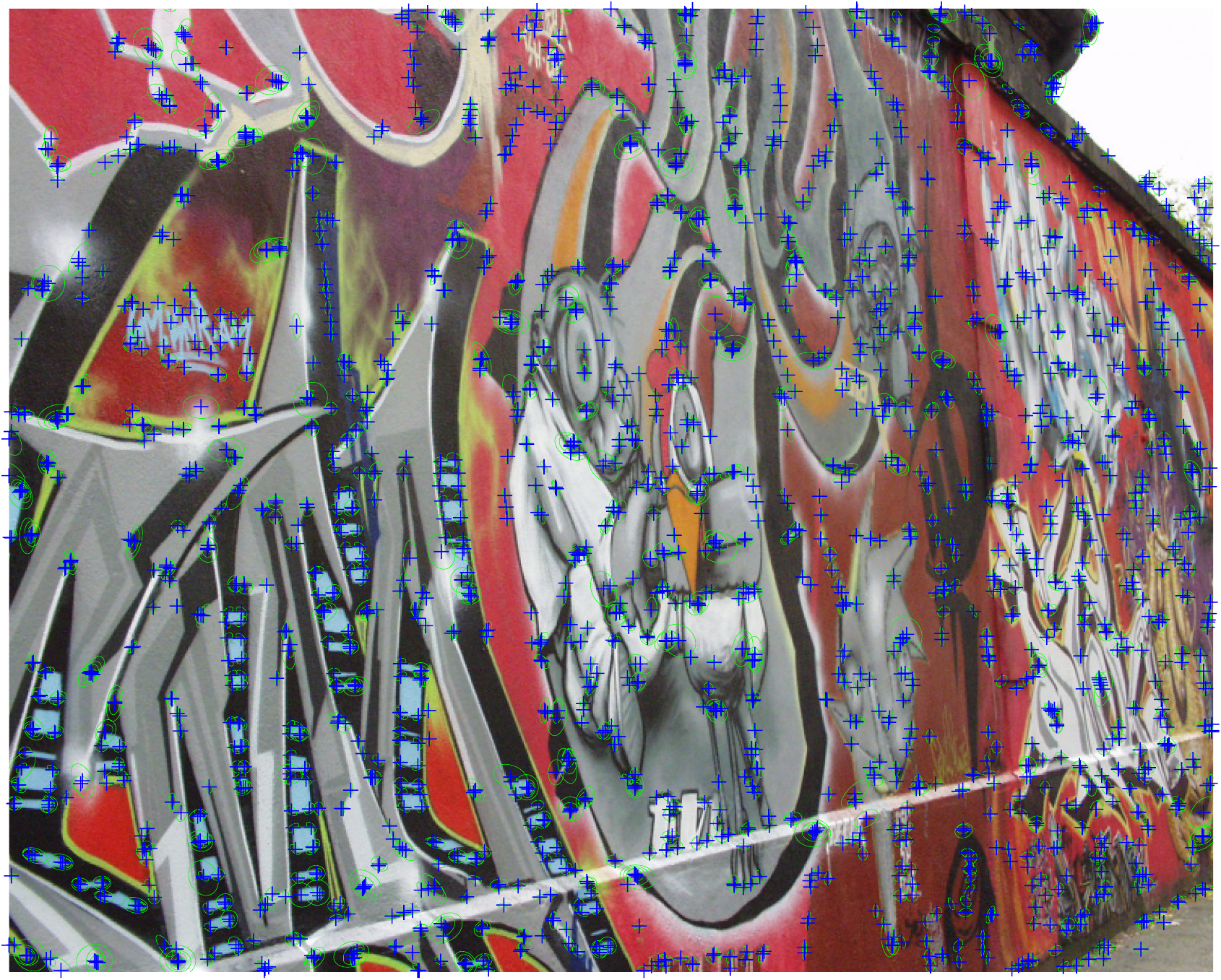} & \includegraphics[height=0.13\textwidth]{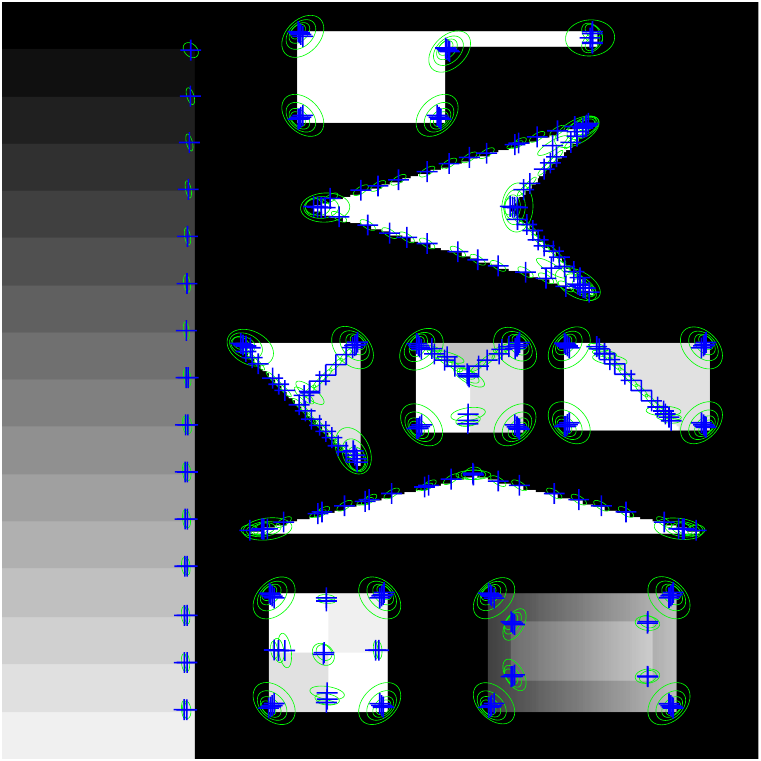} & \includegraphics[height=0.13\textwidth]{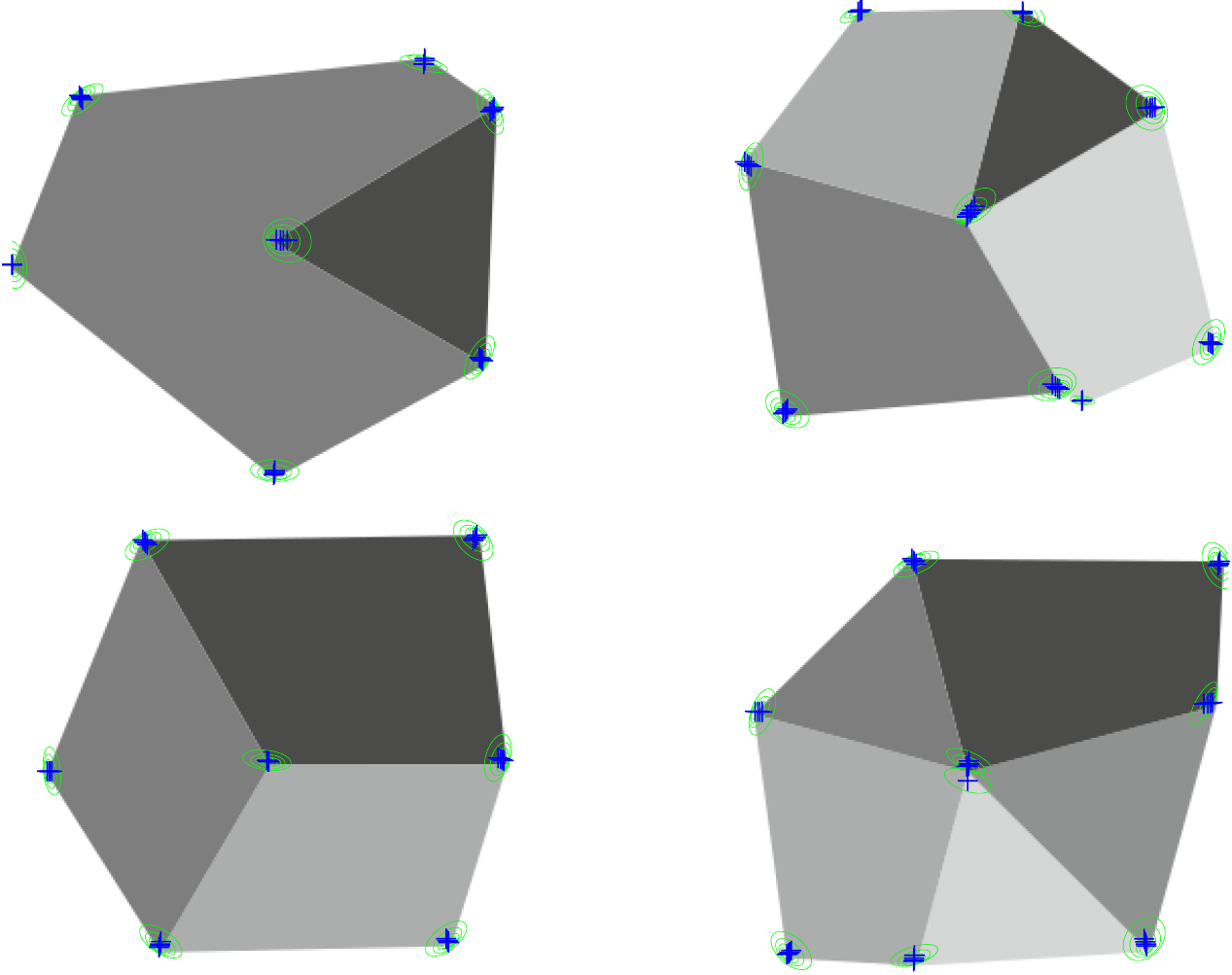}\\
	\end{tabular}
	\caption{\label{hz_hzplus}
		Comparison between HarrisZ and HarrisZ$^+$ on example images, for each keypoint the ellipses associated to the affine transformation determined by $\mu_\sigma$ is shown in green (see text for details, best viewed in color and zoomed in).}
\end{figure}

HarrisZ$^+$ differs from the original HarrisZ by the fine tuning of the parameters in combination with fine implementation changes: the devil is in the details. These updates are strongly related to the advances introduced in the successive steps of the matching pipeline, each one leading to state-of-the-art improvements. In particular:
\begin{enumerate}[i]
    \item\label{affnet}AffNet~\citep{affnet} provides a better patch normalization than the canonical one of SIFT, which gives only the scale\footnote{patch orientation is not taken into account since only almost aligned images are considered, see Sec.\ref{evaluation}}, and it also performs better with respect to the affine transformation estimated by the autocorrelation matrix~\citep{hess_lapl_affine}.
    \item\label{hardnet8}The last version of HardNet~\citep{hardnet8} provides a more discriminant and robust keypoint descriptor than SIFT.
    \item\label{dtm}Blob matching strategy with the inclusion of spatial constraints based on the Delaunay Triangulation Matching (DTM,~\cite{blob_dtm}) retains more correct matches while discarding wrong matches better than the simple Nearest Neighbour Ratio (NNR) strategy~\citep{sift}.
    \item\label{adalam}Filtering matches by multiple local RANSAC has proven to be useful in removing wrong matches with the Adaptive Locally-Affine Matching (AdaLAM,~\cite{adalam}).
    \item\label{degensac}The Degenerate SAmple Consensus (DegenSAC,~\citep{degensac}) checks for degenerate model configurations in RANSAC and obtains better results than the standalone RANSAC. 
\end{enumerate}
Furthermore, the computational power and memory availability of recent hardware greatly increased in the last decade. On this basis, HarrisZ$^+$ adds incrementally the following updates and upgrades with respect to HarrisZ:
\begin{enumerate}
	\item\label{s1} In the original HarrisZ, keypoints are scored first by their decreasing scale indexes $i$ and then by $H_\sigma$ so as to extract the most robust keypoint since higher scales imply wider patch support regions. Nevertheless, the keypoint localization accuracy at high scales degrades. The potential loss in the discriminative power of the local keypoint can be compensated by the choice of a better keypoint descriptor, so HarrisZ$^+$ scores keypoints first by $H_\sigma$ and then by the decreasing scale index $i$. Furthermore, the First Geometric Inconsistent Nearest Neighbor (FGINN,~\cite{mods}) is employed instead of NNR for matching, since FGINN can handle better~\cite{imw2020} than NNR matches in case of spatially close keypoints, which are promoted by the new keypoint ranking. Specifically, if a keypoint $k_1$ on an image $I_1$ matches with a keypoint $k_2$ on the other image $I_2$, this means that their descriptor distance is the minimum $d$ among the distances of remaining keypoints in $I_2$ with $k_1$. In order to filter out unreliable matches, NNR normalizes $d$ by the minimum distance $d'$ between $k_1$ and the remaining keypoints in $I_2$ excluding $k_2$ and thresholds this ratio at a value typically in $[0.8,0.95]$. This greatly increases precision among the tentative matches, but often reduces recall by filtering out correct matches. FGINN restricts the minimum $d'$ only to keypoints far more than 10 px with respect to $k_2$, so closely detected points do not hurt each others.
	\item\label{s2} For ranking keypoints in the case where a constraint on their maximum number has been imposed, a strategy inspired by the greedy local maxima selection of Sec.~\ref{hz_desc} is employed to better distribute the corners over the images, avoiding image regions with clusters of keypoints which can reasonably be associated to the same true corners. If the input image is of size $m\times n$ px and a constraint of having no more than $k$ keypoints is imposed, the minimum required distance $q$ between two keypoints is set as the diameter of a circle with area $\frac{mn}{k/2}$, i.e.
	\vspace{-0.5em}
	\begin{equation}
		q={(2^3mn)}^{\frac{1}{2}}/{\pi k}
	\end{equation}
	Keypoints are then sorted as indicated above, and the greedy local maxima selection using $q$ as distance threshold is performed. If less than $k$ keypoints survive, the process is repeated again on the discarded keypoints. Notice that only roughly $\frac{k}{2}$ keypoints are expected to be selected in the first iteration, so it is more likely that in the second iteration some keypoints close to those found in the first iteration will be chosen. According to preliminary evaluations, this strategy leads to better matching results, probably due to the fact of allowing to some extent closer keypoint, which  generally have some variation within their descriptors, introduces a further chance to match the right descriptors.
	\item\label{s3} The scale index $i$ for $\sigma$ in Eq.~\ref{scale_idx} is set to $0\le i\le4$. The original scale index range $3\le i\le8$ was chosen to limit to about 2000 keypoints or less the detector output for common input image resolution (1024$\times$768 px), but also to extract the most robust keypoints, since higher scales imply wider patch support regions, but at the expense of the keypoint localization accuracy. The new range can extract more than 8000 keypoints, which according to recent evaluations greatly increase the matching power~\citep{imw2020}. Moreover, at finer scale the keypoint localization accuracy is improved. Again, the potential loss in the of discriminative power of the patches is compensated by the choice a better keypoint descriptor.
	\item\label{s4} For color images, the original HarrisZ detector uses the common approach of performing a grayscale conversion of the input using only
	the luminance channel~\citep{gonzalez}. Another popular choice for graylevel conversion is the use of the value channel of the HSV decomposition, corresponding to take the maximum value over the RGB channels. This further grayscale conversion can highlight different image structures, including edges. According to this observation, the edge mask $M_\sigma$ in Eq.~\ref{mask} is computed from further enhanced gradient derivatives $I^\star_x$ and $I^\star_y$ of $I$ which take into account both the luminance $L$ and value $V$ channels in the grayscale conversion of the input image $I$. Specifically, the base horizontal derivatives of $L$ and $V$ are computed as for Eq.~\ref{dx}
	\begin{equation}
		L_x=L\ast K \qquad\text{and}\qquad V_x=V\ast K
	\end{equation}
	so that the maximum absolute gradient value at each pixel location $\mathbf{x}$ is chosen
	\begin{equation}
		I^\star_x(\mathbf{x})=\left\lbrace\begin{matrix}
			L_x(\mathbf{x}) & \text{if}\;\vert L_x(\mathbf{x})\vert > \vert V_x(\mathbf{x})\vert \\
			V_x(\mathbf{x}) & \text{otherwise}
		\end{matrix}\right.
	\end{equation}
	and likewise for $I^\star_y$ from the vertical derivatives $L_y$ and $V_y$. As shown in Fig.~\ref{hz_mask}, $I^\star_x$ and $I^\star_y$ produce a better edge mask $M_\sigma$ but, according to our preliminary experiments, they should not be used for computing the filter response $H_\sigma$ since they lead to worse results.
	\begin{figure}[ht]
		\center
		\vspace{-1em}
		\subfloat[$I$\label{input_img}]{
			\includegraphics[height=0.13\textwidth]{imgs/graf5.jpg}
		}
		\subfloat[$M_{\sqrt{2}}$ differences\label{mask_diff}]{
			\includegraphics[height=0.13\textwidth]{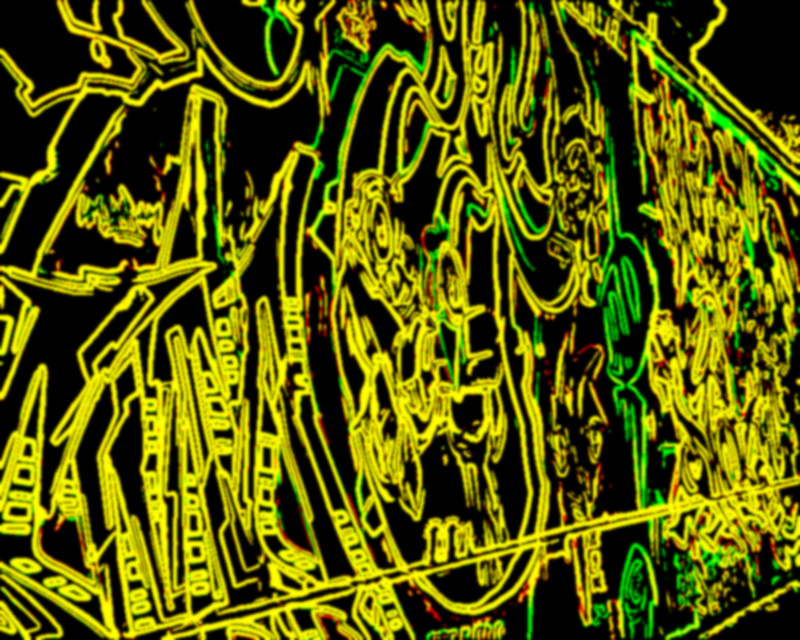}
		}
		\caption{\label{hz_mask}
			Computation of the mask $M_\sigma$ for a color image $I$. Mask portions retained in any case are in yellow, while mask regions considered only with or without the integration of the HSV value channel are respectively in green and red (see text for details, best viewed in color and zoomed in).}
	\end{figure}
	\item\label{s5} In the actual setup, keypoint patches at scale index $i=0$ will get a support region of $25\times25$ px, while HardNet descriptor input patches are $32\times32$ px~\citep{hardnet}. Upsampling would be required in this case, which can introduce noise and spurious details affecting the descriptor robustness. In order to avoid patch upsampling, the final scale associated to the keypoints extracted at $i=0$ at the end of the computation is replaced with the scale value associated to $i=1$, which provides $35\times35$ px patches. Patch downsampling does not negatively affect the HardNet descriptor.
	\item\label{s6} As keypoints for $i=0,1$ get the same final scale $\sigma$, i.e.~both scale indexes output $35\times35$ px patches, local maxima on their union is considering following again the greedy approach with a distance of 1 px. This avoids almost-duplicated keypoint patches.
	\item\label{s7} Before being processed as described in Sec.~\ref{hz_desc}, the input image for scales with indexes i $=0,1$ is doubled using Lanczos interpolation so as the scale value $\sigma$ to reflect the image size adjustment. Extracted keypoints are then reported back to the original reference system by halving the coordinate and scale values. This modification aims at introducing sub-pixel quality details and hence to finer characterizing keypoints, also under the observation that discretized Gaussian kernels with standard deviations around 1 pixel for practical purposes provide basically the same convolution output ($\sigma=1,1.4$ px for $i=0,1$). Besides Lanczos, the bilinear and bicubic interpolations were considered, but they provided worse results.	
\end{enumerate}
Figure~\ref{hz_hzplus} shows the differences between HarrisZ and HarrisZ$^+$, underling the relations between scales and localization accuracy, as well as the different distribution of the corners on the image. 

\vspace{-2em}
\section{Evaluation}\label{evaluation}
\vspace{-0.5em}
\subsection{Setup}
The evaluation relies on the IMC and SimLocMatch contests.

The IMC benchmark~\citep{imw2020} evaluates HarrisZ$^+$ as the upstream module of the image matching pipeline steps~\ref{affnet}-\ref{degensac} described in Sec.~\ref{hzplus}, considering the camera pose error obtained on different scenes. In detail, the error is defined in terms of the mean Average Accuracy (mAA), which takes into account the error achieved at the end of the matching pipeline after the stereo or multiview camera pose estimation. mAA is computed by integrating the maximum angular error between the rotation and translation vectors of the final estimated fundamental matrix up to a threshold of $10^\circ$ for each tested image pairs of the scene. For this aim, reliable pseudo ground-truths obtained by supersets of the image sequences are employed. mAA correlates well with the matching score, but not with repeatability, as the latter is sensitive to the number of keypoints and does not generally agree with the desired and expected results. The IMC benchmark datasets are Phototourism, PragueParks and GoogleUrban, all made up of real images and including validation and test sets. Overall, the complexity from one dataset to another is increasing: each Phototourism scene consists of a main large-scale foreground object almost taken in frontal position, PragueParks includes small-scale scenes with multiple complex objects and higher camera pose variations, and GoogleUrban presents challenge middle-scale urban scene environments with a low level of image overlap. The test sets of the IMC datasets contain respectively 10, 3 and 17 image sequences. Except GoogleUrban, where each image sequence contains 64 or 75 images, other datasets include 100 images for each sequence. Besides the three Phototourism validation sequences employed to check and adjust HarrisZ$^+$ steps, the other validation sets were only used to setup the best DegenSAC threshold and not to fine tune any network parameter in the pipeline, unlike other competitors. Lastly, following the IMC protocol, evaluation is done by restricting the maximum number of allowed keypoints to 2048 (2K) and 8000 (8K).

Differently from IMC, the SimLocMatch benchmark relies on rendered images from synthetic scenes. In this way not only the poses are available for validating the results, but also the exact depth maps so that the evaluation can be directly focused on matches instead of using indirectly the camera pose estimation. The SimLocMatch dataset includes seven scenes representing man-made outdoor and indoor environments for a total of about 10000 images, with different simulated lighting conditions. In the evaluation, matches from roughly 80000 image pairs are checked and, unlike IMC, these include negative image pairs, i.e. with no matches between the images. As evaluation criterion, the SimLocMatch benchmark considers the matching score and the number of wrong matches per image pair, respectively having higher and lower values in the case of better methods.

Notice that both benchmarks only consider almost upright images, i.e. with no relevant relative rotation, which is the most common user application scenario and represents the standard training data for deep architecture. As rotation invariance decrease matchability, the former is disabled for a fair comparison in all methods.  

The Matlab code for both HarrisZ and HarrisZ$^+$ corner detectors is freely available to download, as well as the full image matching pipeline implementation\footnote{\scriptsize{\url{https://sites.google.com/view/fbellavia/}}} which makes use of the code available from the respective authors or through the Kornia library~\citep{kornia}.

\vspace{-1em}
\subsection{Results}
Table~\ref{harrisz_validation} shows the results on the IMC2021 Phototourism validation set obtained by incrementally incorporating the additions presented in Sec.~\ref{hzplus} to get HarrisZ$^+$ from HarrisZ, the final HarrisZ$^+$ corresponds to the last row. For this evaluation, the successive steps of the pipeline consider the AffNet for patch extraction, HardNet8 for computing the descriptors and DegenSAC to estimate the pose (listed as~\ref{affnet},~\ref{hardnet8} and~\ref{degensac} in Sec.~\ref{hzplus}). Spatial local filtering (\ref{dtm} and~\ref{adalam}) is excluded to better evaluate the keypoint extraction process. This ``ablation study'' only considers the stereo setup, excluding multiview pose optimization by bundle adjustment with COLMAP~\citep{colmap}. Each addition from HarrisZ can be applied independently with moderate improvements, however only when taken altogether the proposed changes get an overall significant gain. Specifically, according to the results, better localized keypoints provided by lower scales (listed as~\ref{s1} and~\ref{s3} in Sec.~\ref{hzplus}) are more critical in 8K setup. On the contrary, when the keypoint budget is reduced to 2K, providing an uniform distribution of keypoints among the image (\ref{s2}) becomes relevant. The remaining extensions (\ref{s4}-\ref{s7}) serve mainly as refinements and although each of them only introduces relatively small improvements, the final mAA increase globally of roughly 0.1 and 0.05 respectively for the 2K and 8K setups.
\vspace{-0.5em}

\begin{table}[ht!]
	\caption{Incremental additions towards HarrisZ$^+$ and corresponding mAA@$10^\circ$ on IMC2021 Phototourism validation set. The \# column refers to the last change described in Sec.~\ref{hz_desc} sequentially included from the original HarrisZ, the final HarrisZ$^+$ corresponds to the last row (see text for details)}\label{harrisz_validation}
	\centering
	\resizebox{.45\textwidth}{!}{
	\begin{tabular}{lccc}
		\multirow{2}{*}{Incremental additions} & \multirow{2}{*}{\#} & \multicolumn{2}{c}{mAA@$10^\circ$}\\
		\cmidrule(lr){3-4}
		& & 2K & 8K\\
		\toprule
		Original HarrisZ & - & 0.4619 & 0.5980\\
		$H_\sigma$ ranking + FGINN & \ref{s1} & 0.4955 & 0.6209\\
		Uniform distribution of corners & \ref{s2} & 0.5335 & 0.6203\\
		Scale indexes $i=0,...,4$ & \ref{s3} & 0.5410 & 0.6422\\
		Color image mask $M_\sigma$ & \ref{s4} & 0.5390 & 0.6515\\  
		Output scale $\sigma$ adjustment for $i=0$ & \ref{s5} & 0.5493 & 0.6520\\ 
		Duplicates removal for $i=0,1$ & \ref{s6} &  0.5636 & 0.6522\\
		Doubling input image size for $i=0,1$ & \ref{s7} & 0.5681 & 0.6565\\
		\bottomrule
	\end{tabular}
	}
\end{table}

\begin{table*}[ht]
	\caption{mAA@$10^\circ$ results on IMC2021 (see text for details).}\label{harrisz_imc2021}
	\centering
	\resizebox{.84\textwidth}{!}{	
		\begin{tabular}{clcccccccc}
			& \multirow{2}{*}{Method} & \multicolumn{2}{c}{Phototourism} & \multicolumn{2}{c}{PragueParks} & \multicolumn{2}{c}{GoogleUrban} & \multicolumn{2}{c}{Total}\\
			\cmidrule(lr){3-4} \cmidrule(lr){5-6} \cmidrule(lr){7-8} \cmidrule(lr){9-10}
			& & 2K & 8K & 2K & 8K & 2K & 8K & 2K & 8K\\
			\toprule
			\multirow{8}{*}{\rotatebox[origin=c]{90}{Stereo}} & HarrisZ$^+$ + AffNet + HardNet8 & 0.4753 & 0.5578 & 0.6358 & 0.7368 & 0.3252 & 0.3567 & 0.4788 & 0.5504 \\
			& + Blob + DTM & 0.4449 & 0.5515 & 0.6176 & 0.7387 & 0.2981 & 0.3343 & 0.4535 & 0.5415\\
			& + Blob + DTM + AdaLAM & 0.4369 & 0.5521 & 0.5918 & 0.7423 & 0.2963 & 0.3327 & 0.4417 & 0.5424 \\
			\cmidrule(l){2-10}
			& Upright SIFT & 0.3827 & 0.5122 & 0.4136 & 0.5369 & 0.2542 & 0.2693 & 0.3501 & 0.4394 \\	
			& DoG + AffNet + HardNet8 & n/a & 0.5573 & n/a & 0.5977 & n/a & 0.3009 & n/a & 0.4853 \\	
			& DISK & 0.5121 & 0.5583 & 0.4589 & 0.5622 & 0.2763 & 0.3295 & 0.4157 & 0.4833 \\	
			\cmidrule(l){2-10}
			& LoFTR & n/a & 0.6090 & n/a & 0.7546 & n/a & 0.4060 & n/a & 0.5898 \\
			& (SuperPoint and DISK) + SuperGlue & n/a & 0.6397 & n/a & 0.8070 & n/a & 0.4395 & n/a & 0.6285\\
			\midrule[\heavyrulewidth]
			\multirow{8}{*}{\rotatebox[origin=c]{90}{Multiview}} & HarrisZ$^+$ + AffNet + HardNet8 & 0.6786 & 0.7367 & 0.4676 & 0.4729 & 0.1602 & 0.2025 & 0.4354 & 0.4707 \\
			& + Blob + DTM & 0.7046 & 0.7606 & 0.4583 & 0.4843 & 0.1625 & 0.2060 & 0.4418 & 0.4836\\		
			& + Blob + DTM + AdaLAM & 0.7132 & 0.7580 & 0.4618 & 0.4711 & 0.1521 & 0.2084 & 0.4423 & 0.4791 \\
			\cmidrule(l){2-10}	
			& Upright SIFT & 0.5545 & 0.6850 & 0.3608 & 0.4810 & 0.0520 & 0.0902 & 0.3224 & 0.4187 \\  
			& DoG + AffNet + HardNet8 & n/a & 0.7269 & n/a & 0.4670 & n/a & 0.1643 & n/a & 0.4527 \\ 		
			& DISK & 0.7296 & 0.7445 & 0.4357 & 0.4590 & 0.1275 & 0.1833 & 0.4309 & 0.4623 \\
			\cmidrule(l){2-10}
			& LoFTR & n/a & 0.7610 & n/a & 0.4712 & n/a & 0.3023 & n/a & 0.5115 \\
			& (SuperPoint and DISK) + SuperGlue & n/a & 0.7857 & n/a & 0.4988 & n/a & 0.3374 & n/a & 0.5406 \\
			\bottomrule
		\end{tabular}
	}
	\vspace{-1em}
\end{table*}

Table~\ref{harrisz_imc2021} reports the full mAA results for both the stereo and multiview setups of the last IMC2021, which adds two further datasets with respect to the Phototourism one presented in the previous IMC2020. The HarrisZ$^+$ pipeline including also blob matching plus DTM spatial filtering (\ref{dtm} in Sec.~\ref{hzplus}) with or without AdaLAM (\ref{adalam}) is evaluated. All the pipelines include DegenSAC (\ref{degensac}) as final step, followed by bundle adjustment in case of the multiview setup. The standard upright SIFT matching is included for reference, while the pipeline that replaces HarrisZ$^+$ with the unchained SIFT, i.e. the DoG detector, obtained results among the state-of-the-art in the previous IMC2020 together with DISK. The remaining entries obtained the best results in IMC2021 and employ in addition learned matching strategies. Among the non-learned matching approaches, the pipelines including HarrisZ$^+$ provide the best results on each dataset with the exception of Phototourism, were results are almost aligned, with some advantages of DISK on the 2K keypoint restriction. Among the compared handcrafted keypoint detectors, HarrisZ$^+$ obtains the best results followed by DoG and then SIFT, implying the goodness of the proposed approach, while adding spatial filtering to the pipeline seems to only improve the results in the multiview setup. Excluding the GoogleUrban dataset and the Phototourism stereo setup, when comparing the pipelines based on HarrisZ$^+$ with respect to the best ones relying on learned matching strategies, results are quite comparable. The performance differences are mostly noticeable within the most challenging GoogleUrban dataset, and are probably due to the ability of these architectures to employ, more or less implicitly, a coarse-to-fine image matching strategy, missing in other approaches.

\begin{figure}[ht]
	\center
	\vspace{-2em}
	\includegraphics[width=0.5\textwidth]{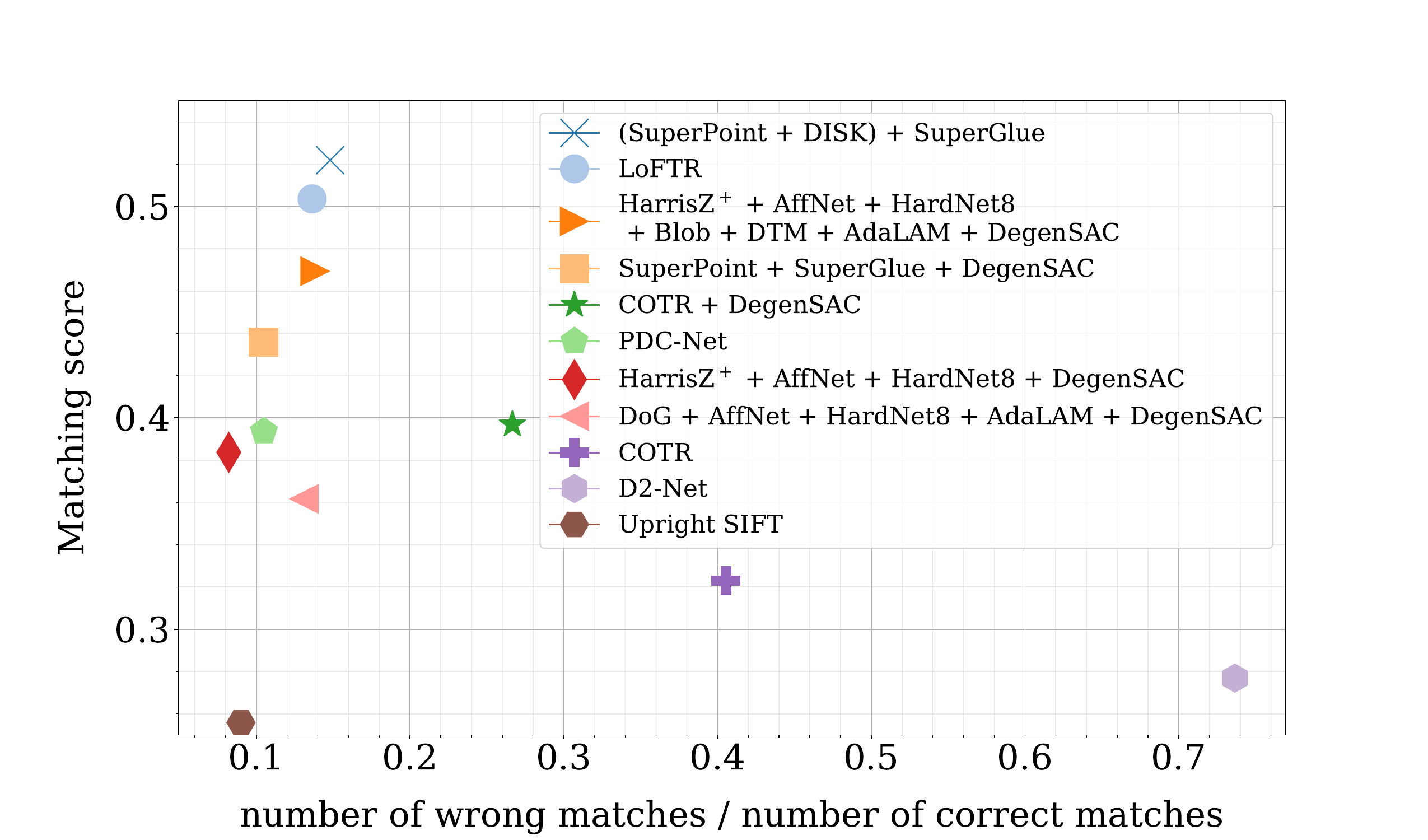}
	\caption{\label{simlocmatch}
		SimLocMatch evaluation results (see text for details, best viewed in color and zoomed in).}
\end{figure}
Figure~\ref{simlocmatch} reports the SimLocMatch results. HarrisZ$^+$ and DoG are limited to 8K keypoints, while the COrrespondence TRansformer (COTR,~\cite{cotr}) and the Probabilistic Dense Correspondence Network (PDC-Net,~\cite{dense}) are included for completeness. Confirming the IMC2021 results, HarriZ$^+$ pipeline completed with spatial local match filtering is the only pipeline able to obtain results very close to those of the approaches employing matching networks based on transformers, i.e. LoFTR and (SuperPoint and Disk) + SuperGlue. In particular, it can be noted that besides achieving a high matching score, also the number of wrongly output matches, normalized by the number of inliers for visualization purposes, is low for HarrisZ$^+$.

Concerning running times, non-optimized HarrisZ$^+$ Matlab code requires on average one second to run on a Intel i9-10900K system with 64 GB RAM and no GPU in case of images with resolutions going from $640\times 480$ to $1900\times 1200$ px, which is roughly five times the original HarrisZ runtime. The bottleneck is the doubled image size for the first two scale indexes. Moreover, HarrisZ$^+$ keypoint uniform ranking is a global process performed at the end of the computation, which can strongly accentuate the computational times in case of a high number of candidate keypoints, as for high-textured images. Nevertheless, these running times are still reasonable for off-line applications.     

As final consideration, classical modular pipeline design, as that HarrisZ$^+$ was designed for, can still offer practical advantages with respect to modern end-to-end deep design for specific applications in terms computational cost or tuning. Adding rotation invariance to the HarrisZ$^+$ pipeline only requires to simply plug OriNet~\citep{affnet} or similar modules before AffNet, while both SuperGlue and LoFTR would require retraining and maybe to add some layers. In any case retrain of a whole end-to-end matching network needs a more expensive system configuration than that mentioned above (with the inclusion of a consumer-grade GPU). On this system, HarrisZ$^+$ pipeline was also tested to work with high resolution images up to $9000\times6732$ px, such those of aerial photogrammetric surveys, while LoFTR was unable to process $2000\times1500$ px input images on a consumer-grade PC either on GPU or CPU.

\vspace{-1.25em}
\section{Conclusions and future work}\label{conclusions}
\vspace{-0.75em}
This paper presents HarrisZ$^+$ corner detector as an upgrade of HarrisZ to be employed with next-gen image matching pipeline. By introducing several changes, which by themselves only slightly affect the final output, HarrisZ$^+$ is able to achieve state-of-the-art results among handcrafted keypoint detectors, providing the basis of competitive image pipelines with respect to the recent end-to-end deep architectures exploiting coarse-to-fine matching strategies thorough transformers. The key idea that guided the HarrisZ$^+$ design is the need of a synergic optimization between the different modules of the pipeline to globally improve the final results, so as to compensate weakness and boost strength points of the whole pipeline.

It is also authors' opinion that research on classical matching methods is still appropriate and effective despite of the overwhelming presence of deep methods. As a further  contribution of this paper, on one hand the results achieved through HarrisZ$^+$ shows that there is still a worthwhile margin of improvement in classic research direction, and on the other hand the gained insight during the development can be employed in turn to design better deep networks as well as to provide more valid training data.

Future work will includes the introduction of handcrafted a coarse-to-fine matching at the end of the pipeline with further optimizations of HarrisZ$^+$ to better adapt to these changes and to improve the computational efficiency, investigations about the possibility to apply HarrisZ$^+$-like selection to DoG keypoints and about using HarrisZ$^+$ keypoints with deep matching architectures based on transformers, such as SuperGlue.

\vspace{-1em}  
\section*{Acknowledgments}
\vspace{-0.5em}  
F. Bellavia is funded by the Italian Ministry of Education and Research (MIUR) under the program PON Ricerca e Innovazione 2014-2020, cofunded by the European Social Fund (ESF), CUP B74I18000220006, id. proposta AIM 1875400, linea di attivit\`{a} 2, Area Cultural Heritage.
D.\ Mishkin is supported by OP VVV funded project CZ.02.1.01/0.0/0.0/$16\_019$/0000765 ``Research Center for Informatics'' and CTU student grant SGS20/171/OHK3/3T/13.
\vspace{-1em}

\bibliographystyle{model2-names}
\bibliography{refs}
\end{document}